\documentclass{article}
\usepackage{spconf,amsmath,graphicx}
\usepackage[skip=0pt]{caption}
\usepackage{amssymb}
\usepackage{multirow}
\usepackage{multicol}
\usepackage{makecell}
\usepackage{color}
\usepackage{cite}
\usepackage[skip=0pt]{caption}

\title{Reflection Removal Using\\Recurrent Polarization-to-Polarization Network}

\name{Wenjiao Bian,  Yusuke Monno, and Masatoshi Okutomi}
\address{Tokyo Institute of Technology,
Tokyo, Japan}

\begin{document}
\ninept
\maketitle
\begin{abstract}

This paper addresses reflection removal, which is the task of separating reflection components from a captured image and deriving the image with only transmission components. Considering that the existence of the reflection changes the polarization state of a scene, some existing methods have exploited polarized images for reflection removal. While these methods apply polarized images as the inputs, they predict the reflection and the transmission directly as non-polarized intensity images. In contrast, we propose a polarization-to-polarization approach that applies polarized images as the inputs and predicts “polarized” reflection and transmission images using two sequential networks to facilitate the separation task by utilizing the interrelated polarization information between the reflection and the transmission. We further adopt a recurrent framework, where the predicted reflection and transmission images are used to iteratively refine each other. Experimental results on a public dataset demonstrate that our method outperforms other state-of-the-art methods.
\end{abstract}
\begin{keywords}
Reflection Removal, Polarization Imaging, Recurrent Neural Network
\end{keywords}
%


\section{Introduction}
Reflections caused by semi-reflectors such as glass are commonly seen in daily life. When light passes through semi-reflectors, a camera inevitably captures the reflection and the transmission components at the same time. Nevertheless, most computer vision applications such as object detection, segmentation, and depth estimation assume that each pixel value is derived only from the scene corresponding to the transmission. Therefore, reflection removal is a crucial task to improve the robustness of real-world applications.

\begin{figure}[!tb]
  \begin{center}
      \includegraphics[width=1\columnwidth]{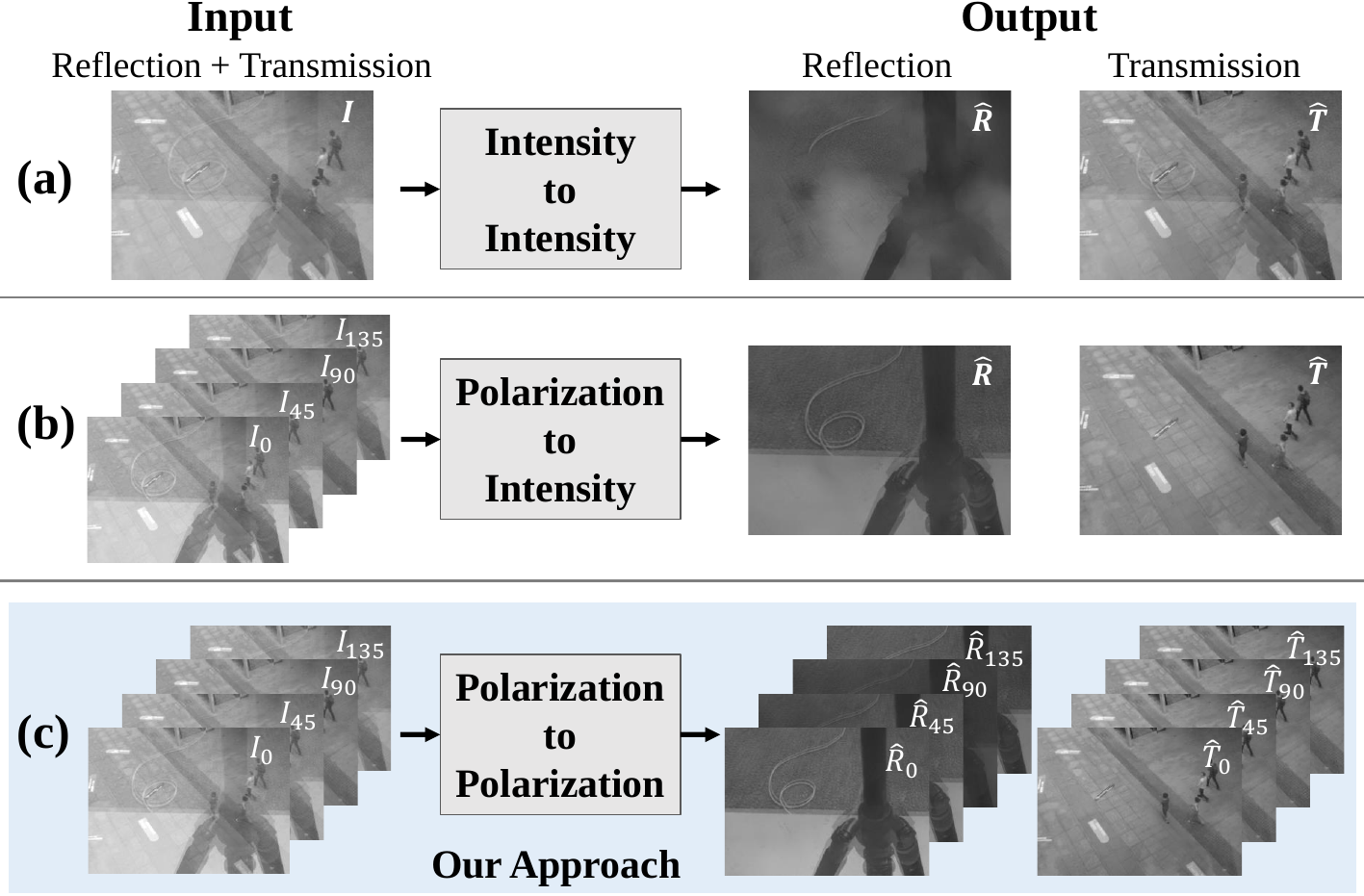}
  \end{center}
  \caption{Different input-output models for the reflection removal. (a)~Both the input and the output of standard single-image methods are intensity images. (b)~Existing polarization-based methods apply polarized images only to the input. (c)~Our proposed polarization-to-polarization approach predicts the output reflection and transmission as polarized images as well.
  }  
  \label{Figure:overview}
\end{figure}

Most existing reflection removal methods are based on a single grayscale or color image, where both the input and the outputs (reflection and transmission) are an intensity domain, as illustrated by the intensity-to-intensity model of Fig.~\ref{Figure:overview}(a). While recent deep-learning-based methods have shown great progress~\cite{ceilnet2017,zhang2018,bdn2018,err2019,ibcln2020,locationsirr2020}, the separation of the reflection and the transmission is still challenging due to an ill-posed problem that an infinite number of the transmission and the reflection image combinations is possible to reproduce the same mixed image.
Other approaches attempt to solve this problem by using multi-view color images \cite{guo2014,han2017,zhu2022}. However, these methods typically necessitate image alignment as a pre-processing step, which imposes constraints on their practical application.

Meanwhile, as the price of one-shot polarization cameras has decreased, one-shot acquisition of polarized images has become much easier in recent years~\cite{maruyama20183,morimatsu2021monochrome}. Considering that the existence of reflection components changes the polarization state of a scene, some non-learning-based \cite{farid1999,sche2000,kong2013,aizu2022} or learning-based \cite{reflectnet2018,lyu2019,lei2020} methods solve the reflection removal by using a set of polarized images with different polarizer orientations (typically, four orientations of $0^\circ$, ${45}^\circ$, ${90}^\circ$, and ${135}^\circ$). While these polarization-based methods apply polarized images as the inputs, they predict the reflection and the transmission images directly as non-polarized intensity images, as illustrated by the polarization-to-intensity model of Fig.~\ref{Figure:overview}(b).

\begin{figure*}[!tb]
     \center
      \includegraphics[width=2\columnwidth]{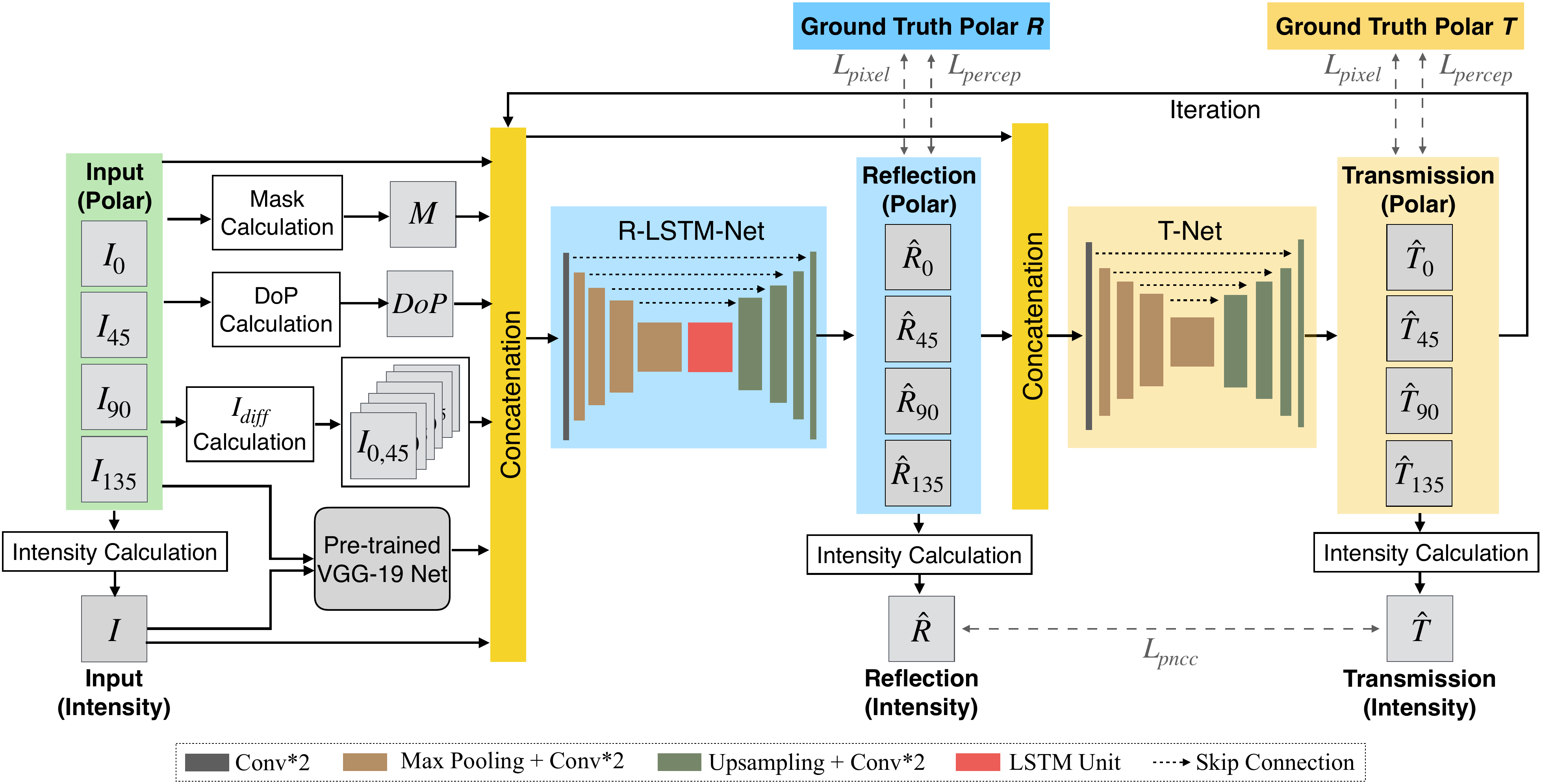}
     \center
  \caption{The overall structure of our proposed RP2PN.}
  \label{Figure:net}
  \vspace{+0.1cm}
\end{figure*}

In this paper, we propose a polarization-to-polarization approach for deep-learning-based reflection removal, as illustrated in Fig.~\ref{Figure:overview}(c). To effectively learn the polarimetric relationships among the input image and the separated reflection and transmission images, our approach takes polarized images as the inputs and predicts the reflection and the transmission images also as the polarized images. Then, the final reflection and transmission outputs are derived as the intensity images by averaging the polarized images.

Regarding the network structure, inspired by~\cite{ibcln2020,locationsirr2020,lei2020}, we propose a two-stage sequential approach within our polarization-to-polarization framework, which uses one recurrent network to predict the reflection and then feeds the reflection result to another network to predict the transmission, as better transmission estimation is also beneficial for reflection estimation and vice versa. We also utilize the difference images between different polarizer angles as the network inputs, because they exhibit an informative feature for the reflection removal.

Experimental results on a public dataset~\cite{lei2020} demonstrate that our method outperforms other state-of-the-art intensity-based and polarization-based methods. Additionally, we highlight the significance of our polarization-to-polarization framework and the effectiveness of the integrated recurrent unit from the ablation study.

\section{Proposed Method}
\subsection{Network Structure}
Figure~\ref{Figure:net} shows the overall structure of our proposed recurrent polarization-to-polarization network (RP2PN), which sequentially and iteratively predicts the polarized reflection and the polarized transmission images. For network training, we use Lei et al. real-world dataset~\cite{lei2020}, which was obtained using Lucid PHX050S-P one-shot monochrome polarization camera equipped with Sony IMX250MZR sensor~\cite{maruyama20183}. This dataset provides the triplets of aligned polarized images $\{I_{\phi}, R_{\phi}, T_{\phi}\}$, where $\phi \in \{0^\circ$, ${45}^\circ$, ${90}^\circ$, ${135}^\circ\}$ is the polarizer angle, $I$ represents the input image with mixed reflection and transmission, and $R$ and $T$ represent the corresponding ground-truth reflection and transmission images, respectively. 

Our RP2PN consists of two sequential networks, namely R-LSTM-Net for the reflection estimation and T-Net for the transmission.  As for the network inputs, from four input polarized images ($I_0$, $I_{45}$, $I_{90}$, $I_{135}$), the intensity image $I$, the degree-of-polarization image ($DoP$) are calculated by a standard polarimetric calculation. 
In addition, we introduce a polarized difference image $I_{diff}$, which serves as informative cues for the reflection removal.

For a mixed polarized image $I_\phi=T_\phi+R_\phi$ captured under a certain polarizer orientation $\phi$, the polarized components of $T_\phi$ and $R_\phi$, denoted as $T^p_\phi$ and $R^p_\phi$, will change with the variation of~$\phi$, while the unpolarized components of $T_\phi$ and $R_\phi$ remain invariant. 
Thus, for two mixed images with the polarizer orientations $\phi_1$ and $\phi_2$, their difference is formed only by the polarized components as
\begin{equation}
\label{eq:I_inten_diff}
    I_{\phi_1}-I_{\phi_2}=T_{\phi_1}^p-T_{\phi_2}^p+(R_{\phi_1}^p-R_{\phi_2}^p).
\end{equation}

\begin{figure}[!tb]
  \begin{center}
      \includegraphics[width=1\columnwidth]{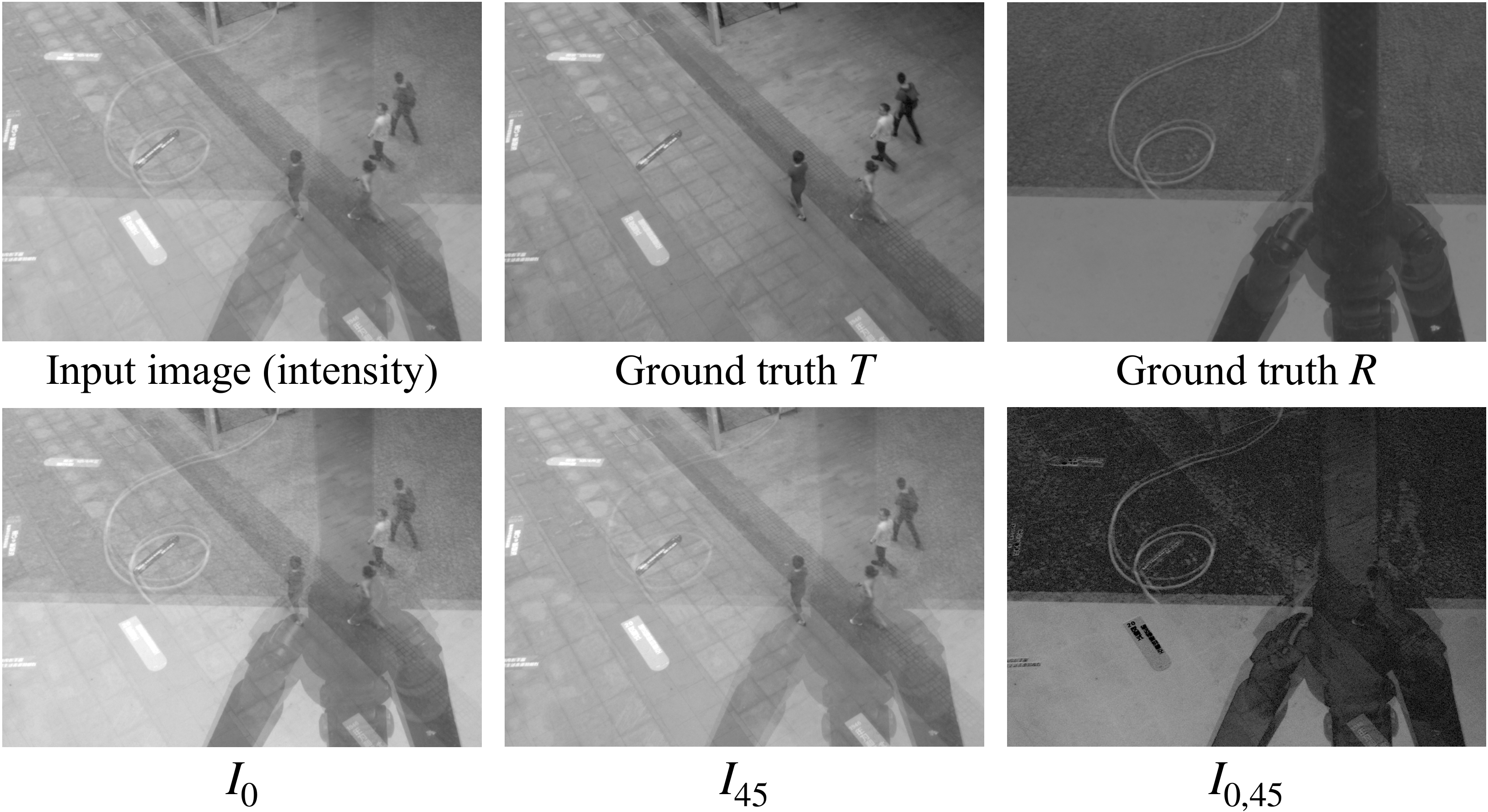}
  \end{center}
  \vspace{+0.1cm}
  \caption{An example of $I_{diff}$ image. $I_{0,45}$ demonstrates closer features to the features of the ground-truth $R$ than either $I_0$ or $I_{45}$. This is due to the polarized $T$ component being considerably weaker than the polarized $R$ component. The brightness of $I_{0,45}$ is adjusted solely for the visualization purpose.}  
  \label{Figure:I_diff}
\end{figure}

We observed in Lei's real-world dataset~\cite{lei2020} that there is a tendency for the strength of polarization (i.e., $DoP$) of the transmission image to be weaker than that of the reflection image. For an example depicted in the first row of Fig.~\ref{Figure:I_diff}, the average $DoP$ values for the ground-truth transmission and reflection are approximately 0.1 and 0.5, respectively. This indicates that the polarized $T$ component is considerably weaker than the polarized $R$ component, resulting in the dominance of the $R$ component in the polarized difference image $I_{diff}$, as shown in the second row of Fig.~\ref{Figure:I_diff}. Based on this observation, we employ all possible combinations of the four polarizer angles to compute the difference images, yielding a total of six polarized difference images ($I_{0,45}$, $I_{0,90}$, $I_{0,135}$, $I_{45,90}$, $I_{45,135}$, $I_{90,135}$), where $I_{\phi_1,\phi_2} = \vert I_{\phi_1} - I_{\phi_2} \vert_1$.

An over-exposure binary mask~($M$) is also derived at each pixel as
\begin{equation}
    \label{eq:OM}
    M = \left\{
        \begin{aligned}
            0,\ \ & \text{if\ \ } max(I_0,I_{45},I_{90},I_{135})>\tau, \\\
            1,\ \ & \text{otherwise}.
        \end{aligned}
        \right.
\end{equation}
where $\tau$ is a threshold and set to 0.98 for the pixel value range of [0,1]. In addition, the features of $I_0$, $I_{45}$, $I_{90}$, $I_{135}$, and~$I$ are respectively extracted from a pre-trained VGG-19 network \cite{vgg19}. All of the above and the original four polarized images are concatenated and fed to the networks as the inputs.

As for our recurrent structure, we first build R-LSTM-Net, which consists of a U-Net architecture~\cite{unet} using a 10-block convolutional encoder and an 8-block decoder with a long short-term memory (LSTM) unit added in the bottleneck \cite{ibcln2020} to predict four polarized reflection images $\hat R_0$, $\hat R_{45}$, $\hat R_{90}$, and $\hat R_{135}$. Then, they are concatenated as a part of the inputs for T-Net with similar U-Net architecture as R-LSTM-Net except for the LSTM unit to predict four polarized transmission images $\hat T_0$, $\hat T_{45}$, $\hat T_{90}$, and $\hat T_{135}$. At last, the predicted polarized transmission images are used as the inputs to further refine the reflection result on the next iteration.

\subsection{Loss Functions}
We apply three loss functions to the last iteration's result of our RP2PN. The total loss $L_{total}$ is defined as
\begin{equation}
\label{eq:loss}
L_{total}=\lambda_1 L_{pixel}+\lambda_2 L_{percep}+\lambda_3 L_{pncc},
\end{equation}
where $\lambda_1$, $\lambda_2$ and $\lambda_3$ are the weighting parameters.

$L_{pixel}$ is the pixel-wise $L_1$ loss between the predicted ($\hat R, \hat T$) and the ground-truth  ($R, T$) images to ensure pixel-level similarity. Different from the existing polarization-based methods~\cite{lyu2019,reflectnet2018,lei2020}, which only consider intensity-domain losses, we evaluate the losses for four polarized images of the reflection and the transmission as
\begin{equation}
\label{eq:l_pixel}
    L_{pixel}=\sum_{\phi\in A}\vert R^{M}_\phi-\hat R^{M}_\phi \vert_1+
    \sum_{\phi\in A}\vert T^{M}_\phi-\hat T^{M}_\phi \vert_1,
\end{equation}
where $A=\{0,45,90,135\}$. The superscript $M$ represents a masked image, e.g., $R_{\phi}^{M}=R_{\phi} \circ M$, where $\circ$ is the pixel-wise production.

$L_{percep}$ is the perceptual loss \cite{perceptual2016} to help the networks to learn high-level contextual features. Similar to $L_{pixel}$, we here calculate the losses in the polarized-domain as 
\begin{equation}
\label{eq:l_p}
\begin{aligned}
    L_{percep}= \sum_{\phi\in A}\sum_j^N
\gamma_j \vert w_{V}^j (R_\phi^{M})-w_{V}^j (\hat R_\phi^{M})\vert_1\\\
+ \sum_{\phi\in A}\sum_j^N
\gamma_j \vert w_{V}^j (T_\phi^{M})-w_{V}^j (\hat T_\phi^{M})\vert_1,
\end{aligned}
\end{equation}
where $w^j_V$ expresses the $j$-th layer's feature map from the pre-trained VGG-19 network and $\gamma_j$ is the weighting parameter of the $j$-th layer.

$L_{pncc}$ is the perceptual normalized cross-correlation loss~\cite{lei2020}, which is applied to minimize the correlation between the predicted reflection and transmission images, assuming their independency. This loss is applied to the final intensity output domain as
\begin{equation}
\label{eq:l_PNCC}
    L_{pncc}=\sum_j^N f_{ncc}(w_{V}^j(\hat R^{M}),w_{V}^j( \hat T^{M})),
\end{equation}
where $\hat R$ and $\hat T$ are the intensity images, which are calculated by the average of four polarized images, and $f_{ncc}$ is the operator to calculate the normalized cross-correlation.


\section{Experimental Results}

\subsection{Implementation Details of Our RP2PN}
We used Lei et al. dataset~\cite{lei2020}, which contains 600, 184, and 107 real-scene polarized image triplets $\{I_{\phi}, R_{\phi}, T_{\phi}\}$ for training, validation, and testing, respectively. The weighting parameters in Eq.~(\ref{eq:loss}) were experimentally set as $\{\lambda_1, \lambda_2, \lambda_3\} = \{0.1, 0.1, 6.0\}$. For the VGG-19 features in Eqs.~(\ref{eq:l_p}) and (\ref{eq:l_PNCC}), we adopted the same six layers ($N=6$) and weights for each layer as~\cite{lei2020}. The number of iterations for RP2PN was experimentally set to three. To train RP2PN, the learning rate was set to $1e^{-4}$ at the first 300 epochs with batch size 1. Then, it was reduced to $1e^{-5}$ for additional 50 epochs. The training took 40 hours using one Nvidia Geforce RTX 3080 Ti GPU. 

\begin{table}[!tb]
\caption{Quantitative comparisons on Lei et al. dataset \cite{lei2020}. \textbf{*}~Non-learning-based methods (Implementation from \cite{reflectnet2018}). \textbf{\dag}~Learning-based methods using pre-trained models.}
\vspace{-0.1cm}
\label{tab:eval}
\begin{center}
\resizebox{\linewidth}{!}{%
\Large
\begin{tabular}{c c c c c c c} 
\hline
\multirow{2}{*}{Methods} & \multirow{2}{*}{\makecell[c]{With \\Polar}} & \multirow{2}{*}{\makecell[c]{Train \\Data}}
& \multicolumn{2}{c}{Transmission} & \multicolumn{2}{c}{Reflection} \\ [1.5pt]
& & & PSNR & SSIM & PSNR & SSIM \\ [1.5pt]
\hline
Farid* \cite{farid1999} & Yes &- &25.56 &0.828 &24.79 &0.742 \\ [1.5pt]
Schechner* \cite{sche2000} & Yes &- &24.62 &0.827 &23.94 &0.621 \\ [1.5pt]
\hline
BDN\dag \cite{bdn2018} & No &\cite{bdn2018} &24.09 &0.756 &23.62 &0.692 \\ [1.5pt]
Dong\dag  \cite{locationsirr2020} & No &\cite{locationsirr2020} & 28.30 & 0.864 & 28.79 & 0.659 \\ [1.5pt]
ReflectNet\dag \cite{reflectnet2018} & Yes &\cite{reflectnet2018} &24.76 &0.821 &25.03 &0.715\\ [1.5pt]
Lyu\dag \cite{lyu2019} & Yes &\cite{lyu2019} & 24.82 & 0.820 & 25.06 & 0.737 \\ [1.5pt]
\hline
Zhang \cite{zhang2018} & No &\cite{lei2020} &32.15 &0.919 &32.20 &0.883 \\ [1.5pt]
IBCLN \cite{ibcln2020} & No &\cite{lei2020}  & 32.84 & 0.928 & 32.80 & 0.897 \\ [1.5pt]
Lei \cite{lei2020} & Yes &\cite{lei2020}  & 35.00 & 0.950 & 34.58 & 0.921 \\ [1.5pt]
RP2PN (Ours) & Yes &\cite{lei2020} & \textbf{35.87} & \textbf{0.954} & \textbf{35.63} & \textbf{0.933} \\ [1.5pt]
\hline
\end{tabular}
}
\end{center}
\end{table}

\subsection{Comparison with Other Methods}

Table~\ref{tab:eval} summarizes the quantitative results for the real-world Lei et al. dataset \cite{lei2020}. We categorize the compared methods into three groups: (i) Non-learning-based methods \cite{sche2000,farid1999}, (ii) learning-based methods using pre-trained models because of the lack of training codes~\cite{bdn2018,locationsirr2020,lyu2019,reflectnet2018}, and (iii) learning-based methods re-trained using Lei et al. dataset and provided training codes~\cite{zhang2018,ibcln2020,lei2020}. For the non-polarization-based methods of \cite{bdn2018,locationsirr2020,zhang2018,ibcln2020}, we used a single-channel intensity image (the average of four polarized images) as the input.

Although the direct outputs of our RP2PN are polarized reflection and transmission images, we evaluated the results in the averaged intensity domain to compare RP2PN with other existing methods. Because there are scale differences in the result images from different methods, we also re-scaled the result images of all the methods as $\hat T'_i=\alpha_i \hat T_i$ and $\hat R'_i=\alpha_i \hat R_i$, where $i$ is the scene index in the testing dataset. The scaling factor $\alpha_i$ was determined for each scene and each method as $\alpha_i = \overline{I_i}/\overline{I'_i}$, where $\overline{I_i}$ is the mean pixel value of the input image and $\overline{I'_i}$ is the mean pixel value of the derived mixed image of $I'_i=\hat T_i+\hat R_i$. With this re-scaling based on the same input image's scale, all methods are more fairly compared.   

\begin{figure}[!tb]
  \begin{center}
    \includegraphics[width=1\columnwidth]{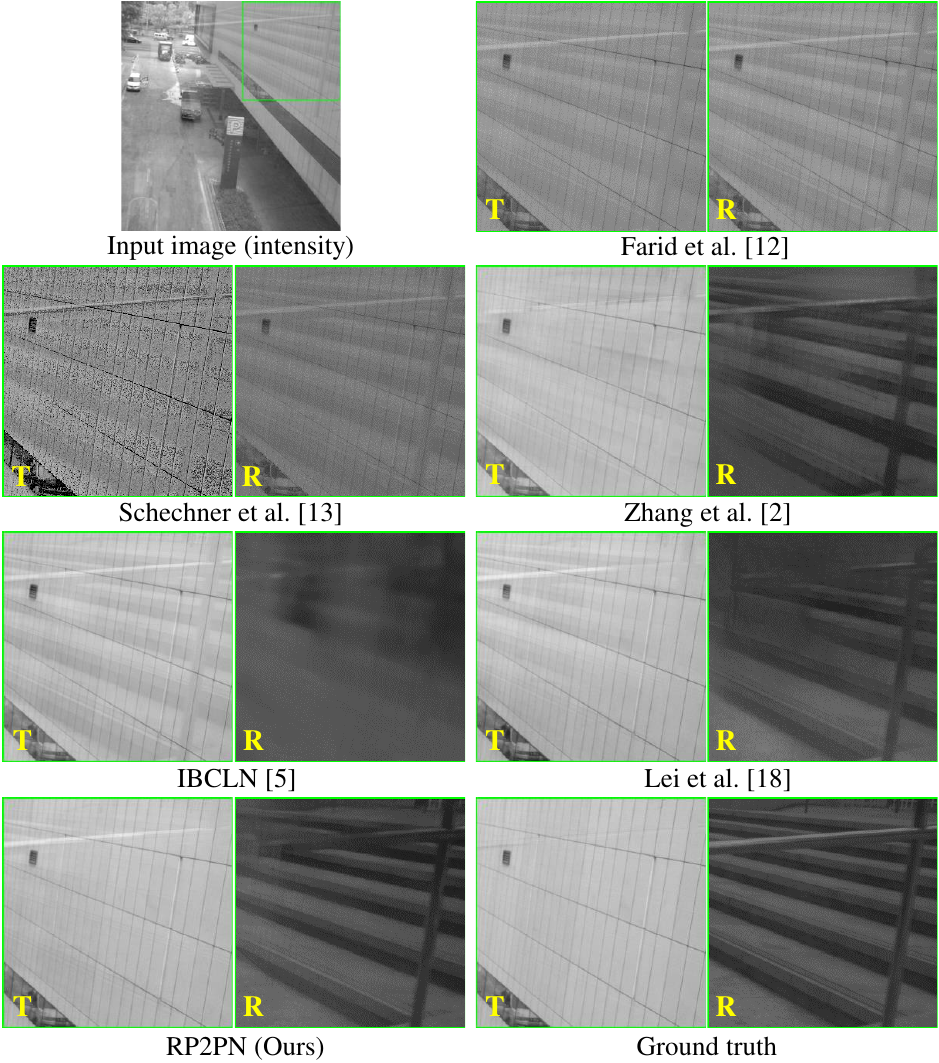}
  \end{center}
  \caption{Qualitative comparison with existing methods.}
  \label{Figure:eval}
  \vspace{+0.5cm}
\end{figure}

\begin{figure}[!tb]
  \begin{center}
      \includegraphics[width=1\columnwidth]{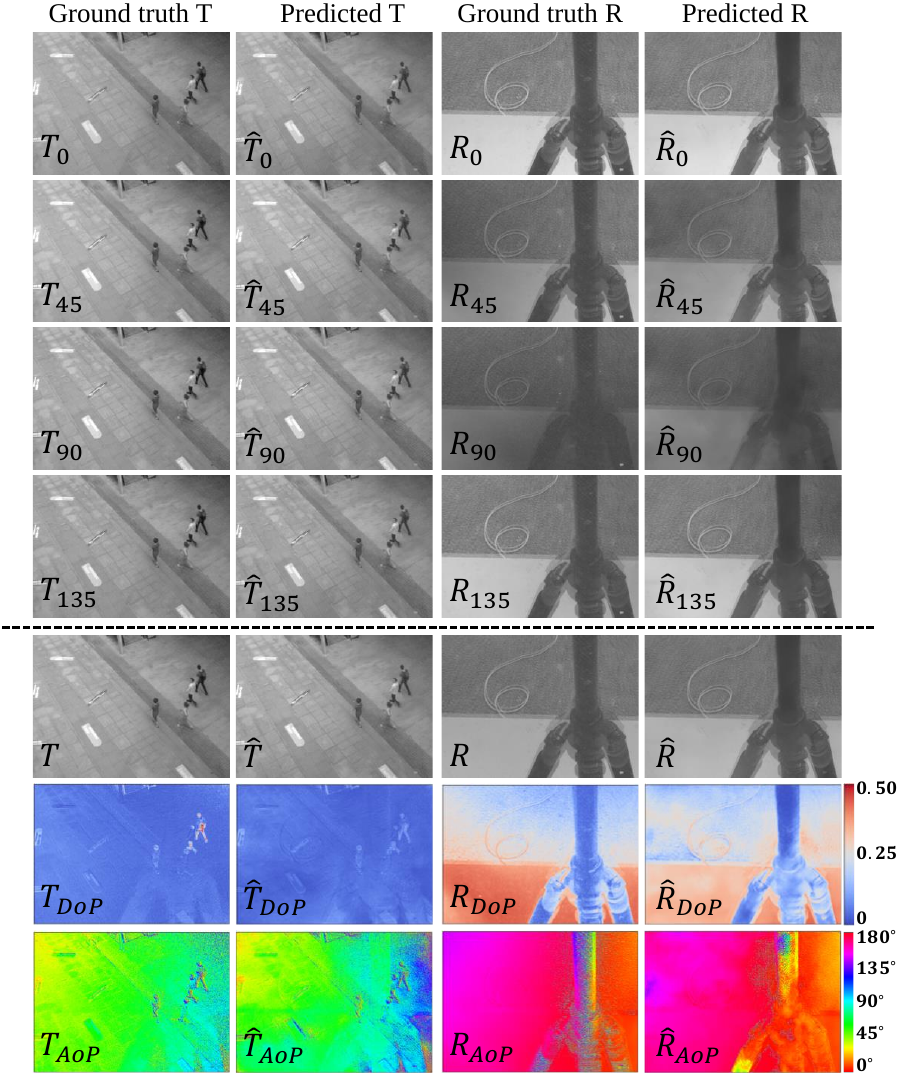}
  \end{center}
  \caption{Example of qualitative results on polarization outputs.}
  \label{Figure:eval_poalr}
  \vspace{+0.2cm}
\end{figure}

The PSNR and SSIM results in Table~\ref{tab:eval} show that non-learning polarization-based methods of~\cite{sche2000,farid1999} exhibit low performance, due to their idealized physical assumptions which are often broken in real-world scenarios. Learning-based polarization methods of~\cite{lyu2019, reflectnet2018} do not achieve the expected performance because the provided pre-trained models were trained on synthetic datasets and showed limited generalizability to Lei et al. dataset. The results of Lei et al. method~\cite{lei2020} and our RP2PN demonstrate higher performance than the non-polarization-based methods of \cite{zhang2018,ibcln2020} using the same training data, which validates the effectiveness of using the polarization. Furthermore, our RP2PN achieves the best PSNR and SSIM results and significant improvement, especially for the reflection. Figure~\ref{Figure:eval} shows the qualitative results (for selected competivie methods due to limited space), where details of each result are shown in green rectangles. Compared with other methods, our transmission result can recover building walls better, while the reflection result preserves the clear edges of the stairs. The results for other scenes can be seen in the supplementary material\footnote{Link: https://github.com/wjbian/RP2PN/blob/main/supp.pdf}. 

Since our RP2PN provides the polarized outputs, we show one example of these outputs in Fig.~\ref{Figure:eval_poalr}. From the results, we can confirm that the polarized reflection and transmission images, as well as the calculated intensity, AoP, and DoP images are reasonably close to the ground truths, which demonstrates that our RP2PN can successfully learn the polarization information of the reflection and the transmission.

\subsection{Ablation Study}
Table~\ref{tab:ablation} summarizes the ablation study results. In models 1 and 2, we replaced the inputs of four polarized images with the standard intensity image to investigate the influence of the polarization input. In models 1 to 3, we replaced the network outputs from four-channel polarized images to one-channel intensity image to investigate the effect of the polarization output. In models 1, 3, and 4, we removed the iteration to investigate the impact of the recurrent framework. Comparing models 1 and 3, utilizing polarized images as the inputs significantly improves the performance. Comparing models 3 and 4, incorporating the polarization output also enhances the separation. Comparing model 4 and ours, it becomes evident that the incorporation of LSTM iterations offers a substantial improvement in predictions, particularly for the reflection. From all of these results, we can confirm that both our polarization-to-polarization approach and recurrent framework are effective.

\begin{table}[!t]
\caption{Ablation study.}
\vspace{-0.2cm}
\label{tab:ablation}
\begin{center}
\resizebox{\linewidth}{!}{%
\Large
\begin{tabular}{c c c c c c c c c}
\hline
\multirow{2}{*}{Model} &\multirow{2}{*}{\makecell[c]{Polar \\Input}} &\multirow{2}{*}{\makecell[c]{Polar \\Output}} &\multirow{2}{*}{\makecell[c]{With \\Iteration}} & \multicolumn{2}{c}{Transmission} & \multicolumn{2}{c}{Reflection} \\  [1.5pt]
& & & & PSNR & SSIM & PSNR & SSIM \\ [1.5pt]
\hline
1  &No  &No &No  &32.93 &0.928 &32.52 &0.892 \\ [1.5pt]
2  &No &No &Yes  &32.97 &0.928 &32.87 &0.897 \\ [1.5pt]
3  &Yes &No &No  &34.95 &0.949 &34.56 &0.921 \\ [1.5pt]
4  &Yes &Yes & No & 35.14 & 0.950 & 34.78 & 0.923 \\ [1.5pt]
\hline
Ours  &Yes &Yes &Yes & \textbf{35.87} & \textbf{0.954} & \textbf{35.63} & \textbf{0.933} \\ [1.5pt]
\hline
\end{tabular}
}
\end{center}
\vspace{-0.5cm}
\end{table}

\section{Conclusion}
In this paper, we have proposed a novel recurrent polarization-to-polarization network, named RP2PN, for reflection removal. Compared with existing polarization-to-intensity approaches, our RP2PN can better utilize the mutual polarimetric relationship between the reflection and the transmission by learning the polarized outputs and incorporating a recurrent framework. The quantitative and qualitative results have validated that our RP2PN is superior to other state-of-the-art methods.

\bibliographystyle{IEEEbib}
\bibliography{bibs/prrnet,bibs/sirr,bibs/polar,bibs/dl, bibs/zadd}

\begin{thebibliography}{10}

\bibitem{ceilnet2017}
Qingnan Fan, Jiaolong Yang, Gang Hua, Baoquan Chen, and David Wipf,
\newblock ``A generic deep architecture for single image reflection removal and image smoothing,''
\newblock in {\em Proceedings of IEEE International Conference on Computer Vision (ICCV)}, 2017, pp. 3238--3247.

\bibitem{zhang2018}
Xuaner Zhang, Ren Ng, and Qifeng Chen,
\newblock ``Single image reflection separation with perceptual losses,''
\newblock in {\em Proceedings of IEEE Conference on Computer Vision and Pattern Recognition (CVPR)}, 2018, pp. 4786--4794.

\bibitem{bdn2018}
Jie Yang, Dong Gong, Lingqiao Liu, and Qinfeng Shi,
\newblock ``Seeing deeply and bidirectionally: A deep learning approach for single image reflection removal,''
\newblock in {\em Proceedings of European Conference on Computer Vision (ECCV)}, 2018, pp. 654--669.

\bibitem{err2019}
Kaixuan Wei, Jiaolong Yang, Ying Fu, Wipf David, and Hua Huang,
\newblock ``Single image reflection removal exploiting misaligned training data and network enhancements,''
\newblock in {\em Proceedings of IEEE Conference on Computer Vision and Pattern Recognition (CVPR)}, 2019, pp. 8170--8179.

\bibitem{ibcln2020}
Chao Li, Yixiao Yang, Kun He, Stephen Lin, and John~E Hopcroft,
\newblock ``Single image reflection removal through cascaded refinement,''
\newblock in {\em Proceedings of IEEE Conference on Computer Vision and Pattern Recognition (CVPR)}, 2020, pp. 3565--3574.

\bibitem{locationsirr2020}
Zheng Dong, Ke~Xu, Yin Yang, Hujun Bao, Weiwei Xu, and Rynson~WH Lau,
\newblock ``Location-aware single image reflection removal,''
\newblock in {\em Proceedings of IEEE International Conference on Computer Vision (ICCV)}, 2021, pp. 5017--5026.

\bibitem{guo2014}
Xiaojie Guo, Xiaochun Cao, and Yi~Ma,
\newblock ``Robust separation of reflection from multiple images,''
\newblock in {\em Proceedings of IEEE Conference on Computer Vision and Pattern Recognition (CVPR)}, 2014, pp. 2187--2194.

\bibitem{han2017}
Byeong-Ju Han and Jae-Young Sim,
\newblock ``Reflection removal using low-rank matrix completion,''
\newblock in {\em Proceedings of IEEE Conference on Computer Vision and Pattern Recognition (CVPR)}, 2017, pp. 5438--5446.

\bibitem{zhu2022}
Chengxuan Zhu, Renjie Wan, and Boxin Shi,
\newblock ``Neural transmitted radiance fields,''
\newblock in {\em Proceedings of Advances in Neural Information Processing Systems (NeurIPS)}, 2022, vol.~35, pp. 38994--39006.

\bibitem{maruyama20183}
Yasushi Maruyama, Takashi Terada, Tomohiro Yamazaki, Yusuke Uesaka, Motoaki Nakamura, Yoshihisa Matoba, Kenta Komori, Yoshiyuki Ohba, Shinichi Arakawa, Yasutaka Hirasawa, Yuhi Kondo, Jun Murayama, Kentaro Akiyama, Yusuke Oike, Shuzo Sato, and Takayuki Ezaki,
\newblock ``3.2-{MP} back-illuminated polarization image sensor with four-directional air-gap wire grid and 2.5-$\mu$m pixels,''
\newblock {\em IEEE Transactions on Electron Devices}, vol. 65, no. 6, pp. 2544--2551, 2018.

\bibitem{morimatsu2021monochrome}
Miki Morimatsu, Yusuke Monno, Masayuki Tanaka, and Masatoshi Okutomi,
\newblock ``Monochrome and color polarization demosaicking based on intensity-guided residual interpolation,''
\newblock {\em IEEE Sensors Journal}, vol. 21, no. 23, pp. 26985--26996, 2021.

\bibitem{farid1999}
Hany Farid and Edward~H Adelson,
\newblock ``Separating reflections and lighting using independent components analysis,''
\newblock in {\em Proceedings of IEEE Conference on Computer Vision and Pattern Recognition (CVPR)}, 1999, vol.~1, pp. 262--267.

\bibitem{sche2000}
Yoav~Y Schechner, Joseph Shamir, and Nahum Kiryati,
\newblock ``Polarization and statistical analysis of scenes containing a semireflector,''
\newblock {\em Journal of the Optical Society of America. A}, vol. 17, no. 2, pp. 276--284, 2000.

\bibitem{kong2013}
Naejin Kong, Yu-Wing Tai, and Joseph~S Shin,
\newblock ``A physically-based approach to reflection separation: From physical modeling to constrained optimization,''
\newblock {\em IEEE Transactions on Pattern Analysis and Machine Intelligence}, vol. 36, no. 2, pp. 209--221, 2013.

\bibitem{aizu2022}
Takuma Aizu and Ryo Matsuoka,
\newblock ``Reflection removal using multiple polarized images with different exposure times,''
\newblock in {\em Proceedings of European Signal Processing Conference (EUSIPCO)}, 2022, pp. 498--502.

\bibitem{reflectnet2018}
Patrick Wieschollek, Orazio Gallo, Jinwei Gu, and Jan Kautz,
\newblock ``Separating reflection and transmission images in the wild,''
\newblock in {\em Proceedings of the European Conference on Computer Vision (ECCV)}, 2018, pp. 90--105.

\bibitem{lyu2019}
Youwei Lyu, Zhaopeng Cui, Si~Li, Marc Pollefeys, and Boxin Shi,
\newblock ``Reflection separation using a pair of unpolarized and polarized images,''
\newblock in {\em Proceedings of Advances in Neural Information Processing Systems (NeurIPS)}, 2019, pp. 14559--14569.

\bibitem{lei2020}
Chenyang Lei, Xuhua Huang, Mengdi Zhang, Qiong Yan, Wenxiu Sun, and Qifeng Chen,
\newblock ``Polarized reflection removal with perfect alignment in the wild,''
\newblock in {\em Proceedings of IEEE Conference on Computer Vision and Pattern Recognition (CVPR)}, 2020, pp. 1747--1755.

\bibitem{vgg19}
Karen Simonyan and Andrew Zisserman,
\newblock ``Very deep convolutional networks for large-scale image recognition,''
\newblock {\em arXiv preprint:1409.1556}, 2014.

\bibitem{unet}
Olaf Ronneberger, Philipp Fischer, and Thomas Brox,
\newblock ``U-net: Convolutional networks for biomedical image segmentation,''
\newblock in {\em Proceedings of Medical Image Computing and Computer-Assisted Intervention (MICCAI)}, 2015, pp. 234--241.

\bibitem{perceptual2016}
Justin Johnson, Alexandre Alahi, and Li~Fei-Fei,
\newblock ``Perceptual losses for real-time style transfer and super-resolution,''
\newblock in {\em Proceedings of the European Conference on Computer Vision (ECCV)}, 2016, pp. 694--711.

\end{thebibliography}
\end{document}